# Arrhythmia Classification from 12-Lead ECG Signals Using Convolutional and Transformer-Based Deep Learning Models

Andrei Apostol, Maria Nuțu

*aDepartment of Mathematics and Informatics, Transilvania University, Brașov, 500036, Romania*

**Abstract**

In Romania, cardiovascular problems are the leading cause of death, accounting for nearly one-third of annual fatalities. The severity of this situation calls for innovative diagnosis method for cardiovascular diseases. This article aims to explore efficient, light-weight and rapid methods for arrhythmia diagnosis, in resource-constrained healthcare settings. Due to the lack of Romanian public medical data, we trained our systems using international public datasets, having in mind that the ECG signals are the same regardless the patients' nationality. Within this purpose, we combined multiple datasets, usually used in the field of arrhythmias classification: PTB-XL electrocardiography dataset [41], PTB Diagnostic ECG Database [3], China 12-Lead ECG Challenge Database [24], Georgia 12-Lead ECG Challenge Database [13], and St. Petersburg INCART 12-lead Arrhythmia Database [37]. For the input data, we employed ECG signal processing methods, specifically a variant of the Pan-Tompkins algorithm, useful in arrhythmia classification because it provides a robust and efficient method for detecting QRS complexes in ECG signals. Additionally, we used machine learning techniques, widely used for the task of classification, including convolutional neural networks (1D CNNs, 2D CNNs, ResNet) and Vision Transformers (ViTs). The systems were evaluated in terms of accuracy and F1 score. We annalysed our dataset from two perspectives. First, we fed the systems with the ECG signals and the GRU-based 1D CNN model achieved the highest accuracy of 93.4% among all the tested architectures. Secondly, we transformed ECG signals into images and the CNN2D model achieved an accuracy of 92.16%.

*Keywords:*

## 1. Introduction

Arrhythmia diagnosing is one of key roles of an electrocardiogram (ECG), a better method for this task is yet to be found. The ECG (illustrated in Figure 1) is an essential instrument in cardiology for assessing a patient's heart condition. Fundamentally, the ECG is a capture of the heart muscle's electrical activity (Figure 2). Changes in the heart's electrical potential are detected on the body's surface using electrodes [26]. The aim of this paper is to develop a computer-aided system to help cardiologists by providing a smart, cost-effective, and time-saving diagnostic solution. To achieve this goal, traditional ECG signal processing techniques are used alongside deep learning methods.

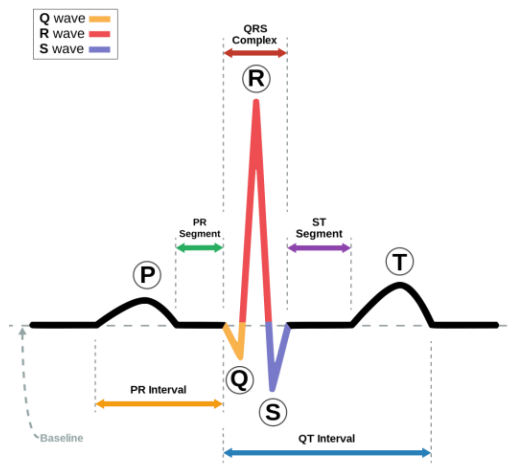

**Figure 1:** Normal ECG waves

Traditional arrhythmia classification methods rely on manually designed features such as heart rate variability and QRS duration. Although these features can provide useful insights, they often fail to generalize across diverse patient populations due to variations in ECG morphology, noise, and recording conditions. In addition, many conventional approaches depend on segmenting individual heartbeats for classification, which can lead to the loss of important temporal patterns that span multiple cardiac cycles. These limitations make rule-based and feature-engineered methods less reliable, especially in real-world clinical applications where ECG signals are highly variable. Deep learning addresses these challenges by allowing automated feature extraction from unprocessed ECG data, thus removing the necessity for manual feature engineering. Neural networks, especially convolutional and recurrent structures like CNNs and LSTMs, adeptly capture the spatial and temporal dependencies present in the signal, resulting in more reliable and precise classification. Compared to traditional approaches, deep learning models offer superior generalization across diverse patients and recording environments, making them more suitable for large-scale, real-time arrhythmia detection.

From 2000 to 2019, the life expectancy in Romania rose by over four years; however, in 2020, it suffered a brief decline of 1.4 years as a result of the COVID-19 pandemic. Cardiovascular diseases remain the primary cause of death, responsible for more than one-third of fatalities in 2018 [27] and over half in 2020 [28]. The healthcare system's inefficiency has brought the avoidable mortality to the third-highest in the EU [27]. This can be attributed to restricted access to healthcare services linked to income; in 2022, Romania was positioned seventh among nations encountering significant challenges in obtaining medical services by income level [10]. Considering the aspects previously discussed, there was a desire to develop a tool to aid in diagnosing patients with cardiovascular conditions, particularly asymptomatic arrhythmias or those with minimal mortality risk. Deep learning methods enhance diagnostic automation, decreasing both the time needed for interpretation and the reliance on physician involvement, thereby reducing analysis costs. Additionally, this tool can act as a foundational resource for training medical students.

The work of Pranav Rajpurkar [31], Tae Joon Jun [21], Zhaohan Xiong [43], and Tsai-Min Chen [5] significantly influenced the foundation of this study. Building on the ideas they formulated, we experimented with similar architectures to classify different types of arrhythmia. In the development of this work, several datasets for electrocardiograms (ECG) were used. For the purpose of optimizing the training process, these datasets were integrated into a single entity, stored as a .npy file.

ECG-based arrhythmia classification presents several challenges due to the complex and highly variable nature of ECG signals. One of the most significant difficulties is accurate QRS detection, as the shape, amplitude, and duration of QRS complexes can vary significantly across patients and different arrhythmia types. Noise, baseline wander, and signal artifacts further complicate the identification of QRS complexes, making robust detection crucial for reliable classification. To address this challenge, we developed a simplified Pan-Tompkins algorithm that effectively detects QRS complexes while maintaining computational efficiency. This method ensures precise localization of key waveform components, which is essential for distinguishing between different conditions such as Sinus Rhythm (SNR), First-Degree Atrioventricular Block (IAVB), Atrial Fibrillation (AF), Supraventricular Tachycardia (STach), and Sinus Bradycardia (SB). By tailoring the algorithm to these specific arrhythmias, we achieved a balance between computational efficiency and detection accuracy, demonstrating its effectiveness in real-world ECG analysis. We developed a simplified Pan-Tompkins algorithm for efficient QRS detection and designed lightweight neural networks that enable faster, real-time arrhythmia classification while maintaining high accuracy, making them suitable for practical deployment, even in low budget environments.

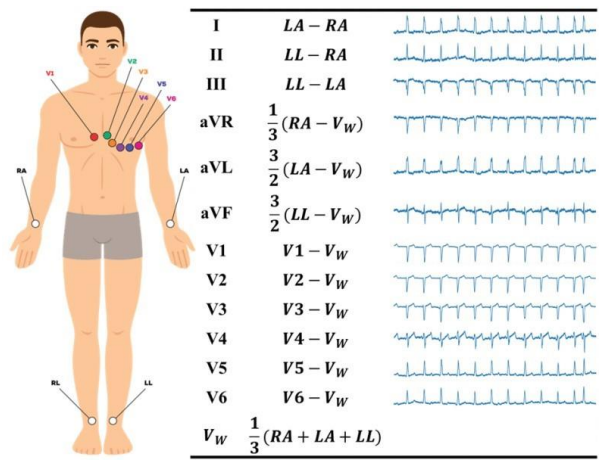

**Figure 2:** 12 lead ECG illustration

In the process of training the model, ECG signals from all 12 leads were used (Figure 2). To achieve better results, a series of features were added, extracted from the ECG signal, representative of various types of arrhythmias, the QRS intervals. These features were obtained through an adapted version of the Pan-Tompkins algorithm, tailored for identifying QRS complexes within the ECG signal.

This paper aims to diagnose cardiovascular diseases, specifically mild, asymptomatic forms, or those with low mortality. To achieve this, we used convolutional neural networks, trained on electrocardiograms corresponding to the types of arrhythmias we aim to diagnose. While other approaches classify individual heartbeats to identify specific arrhythmias, our method analyzes the entire ECG signal, giving particular attention to QRS complexes for improved classification, capturing broader temporal patterns and leveraging QRS features. During the training process, we used the full depiction of the electrical activities, along with critical points of interest that are indicative for recognizing arrhythmias, such as the QRS complexes.

The paper is organised as follows: First, recent related work on arrhythmia classification and ECG signal processing is reviewed in Section 2. We also addressed recent research on medical image classification (Section 2.1). In Section 3, the theoretical background underlying the proposed methodologies is presented. Section 3.1 provides an overview of the arrhythmias studies and their clinical relevance. The experimental setup, including datasets and preprocessing techniques, are detailed in Section 4 and Section 5. The systems' architectures are described in Section 6. The results are presented and discussed in Section 7. Finally, Section 8 concludes the paper by assessing the current state of the proposed method and suggesting future directions for development.

## 2. Related work

The field of ECG signal analysis and arrhythmia classification has evolved significantly in recent years, thanks to advances in machine learning and deep learning. Numerous studies have explored various neural network architectures and preprocessing techniques to improve the accuracy of automated diagnosis.

A key reference for this paper was the article published by Rajpurkar and Hannun [31]. In this study, the authors developed a convolutional neural network (CNN) model capable of detecting arrhythmia at a level comparable to that of cardiologists. The model was trained on a large dataset consisting of 64,121 ECG recordings, covering a duration of 30 seconds each. The neural network was designed to identify 14 different types of cardiac rhythms, including atrial fibrillation and various forms of heart block. The article presents a deep convolutional neural network architecture (CNN) for detecting cardiac arrhythmia directly from raw ECG signals. The architecture consists of several convolutional and pooling layers, followed by dense layers and a softmax output layer, allowing for accurate arrhythmia classification. This comprehensive model autonomously identifies essential features, negating the necessity for manual feature extraction from ECG data. While the article showcases promising outcomes in arrhythmia detection using convolutional neural networks, subsequent implementations have faced difficulties in maintaining similar consistent performance. The F1 score for the proposed method ranges between **0.656** and **0.939**, with an average of approximately **0.809**.

The study of Isin and Ozdalilib [18] uses convolutional neural networks and is notable for employing AlexNet to extract significant features from the R-T intervals of ECG signals. These features are then dimensionally reduced using Principal Component Analysis (PCA), thereby decreasing computational load and reducing the risk of overfitting[1]. This approach leverages deep transfer learning to avoid training networks from scratch, focusing on the automated classification of cardiac conditions such as normal rhythm, incomplete right bundle branch block (RBBB), and paced beats, achieving a test accuracy of approximately 92%.

In the collection of our reviewed articles, the research of Dhyani et al. [8] is noteworthy due to its implementation of the Support Vector Machine (SVM) technique for the analysis and diagnosis of cardiac arrhythmias using electrocardiograms (ECG). Their method showcased impressive accuracy in classifying nine different heartbeat types within the CPSC 2018 dataset[2] 7000 recordings

The study also employed 3D Discrete Wavelet Transform (DWT) for ECG signal preprocessing, encompassing noise reduction and essential feature extraction, resulting in a powerful approach for the analysis and interpretation of complex cardiac signals, with an accuracy reported of around 98%.

Taloba et al. [35] made notable advancements by integrating various methods for detecting and analyzing arrhythmias from ECG signals. Their research introduces a novel approach combining TERMA (Temporal Event Recognition and Model Analysis) with the fractional Fourier transform (FFT). A prominent feature of their work is the employment of a Moving Average (MA) for recognizing ECG signals, including T waves and P, QRS complexes, which, along with the use of TERMA — primarily used in economic market analysis — marks an important development in biomedical signal analysis. The study is structured into four major stages for ECG signal classification: ECG signal preprocessing, QRS complex detection and signal segmentation, parameter extraction, and classification using supervised learning algorithms such as MLP (Multilayer Perceptron) and SVM (Support Vector Machine). By applying these methods, the researchers achieved precise detection of R peaks as well as P and T waves, which are essential for diagnosing cardiac arrhythmias, reaching 89% accuarcy.

---

[1]In machine learning, "overfitting" refers to a model fitting too closely to the training data, which can hinder its ability to generalize well to new data [45]

[2]China Physiological Signal Challenge 2018. (2018). Classification of cardiac arrhythmias from 12-lead ECG recordings.

Tae Joon Jun et al. [21] introduced an innovative technique for classifying arrhythmias utilizing a two-dimensional convolutional neural network (2D CNN). Within this study, the researchers converted one-dimensional ECG signals into two-dimensional images to incorporate both temporal and frequency data. Rather than analyzing the entire ECG signal, the network was trained on segmented images of heartbeats to emphasize the unique characteristics of each beat. The authors employed a 2D CNN that was trained with a diverse dataset encompassing different arrhythmia types, enhancing its ability to generalize. Their methodology incorporates data preprocessing strategies, notably applying the Short-Time Fourier Transform (STFT) to transform ECG signals into a two-dimensional format. This method allows the network to extract intricate spatial characteristics and classify arrhythmic beats with 99% accuracy.

Li et al. [23] proposed a two-dimensional ECG-based approach for cardiac arrhythmia classification using a deep learning model called DSE-ResNet. The DSE-ResNet model integrates depthwise separable convolutions and residual connections to enhance feature extraction while reducing computational complexity. Experimental results demonstrated that this approach significantly improves classification accuracy and generalization across different datasets, making it a promising method for automated arrhythmia detection.

Zhang et al. [47] introduced an interpretable deep learning framework for diagnosing cardiac arrhythmias from 12-lead ECG recordings. Their model employs a convolutional neural network (CNN) to perform multi-label classification, detecting various arrhythmia types with high accuracy. A key aspect of their work is the incorporation of SHapley Additive exPlanations (SHAP), which provides insights into the model's decision-making by highlighting the most relevant ECG features contributing to each prediction.

This interpretability enhances clinical trust in AI-driven diagnostics and aids physicians in understanding automated ECG analysis.

**Table 1:** Literature results for arrhythmia classification

| Reference | Input data | No of classes | Architecture | Accuracy |
| --- | --- | --- | --- | --- |
| Rajpurkar and Hannun [31] | raw ECG signals | 14 cardiac rhythms | CNN | 80% |
| Zhang et al. [47] | 12-lead ECG signals | 9 arrhythmia types | CNN + SHAP | 97.9% |
| Isin and Ozdalilib [18] | R-T intervals of ECG signals | 3 cardiac conditions | AlexNet | 92% |
| Dhyani et al.[8] | CPSC 2018 dataset | 9 heartbeat types | SVM | 99% $_s$ |
| Taloba et al. [35] | Heartbeat features | 2 heartbeat types | SVM | 68% |
|  |  |  | MLP | 99% |
| Tae Joon Jun et al. [21] | Images of each heartbeat type | 8 heartbeat types | CNN | 99% |
| Li et al. [23] | 2D spliced ECG signals | 9 arrhythmia types | DSE-ResNet | 96.4% |

## 2.1. Medical Image Classification

After achieving remarkable success in natural language tasks, Transformers have been effectively adapted to various computer vision challenges, delivering state-of-the-art performance and leading researchers to question the dominance of convolutional neural networks (CNNs) as the default choice. Building on these advancements in computer vision, the medical imaging field has also seen increasing interest in Transformers, which can capture global context more effectively than CNNs, which rely on local receptive fields [33].

The study conducted by Liangrui Pan et al. [30] introduces an innovative framework for classifying images, intended for detecting osteosarcoma in histological images. Osteosarcoma is a cancerous bone tumor that complicates early detection because of its intricate visual patterns in microscopic imagery. Earlier methods, like CNNs and other machine learning techniques, often struggle with noise management and capturing multiscale contextual details. To overcome these limitations, the researchers introduced a hybrid approach incorporating vision transformers (ViTs) with a Noise Reduction Convolutional Autoencoder (NRCA) and a Feature Cross Fusion Learning (FCFL) layer. These elements work together to enhance the precision and reliability of classifying tumors. The NRCA-FCFL implementation starts with noise reduction, using the NRCA to preprocess histological images by filtering out irrelevant noise. The images are then divided into patches at different scales, where separate ViTs process these patches to derive multiscale features. An evaluation conducted on a publicly available dataset for osteosarcoma [22] demonstrated an impressive accuracy of **99.17%**

The paper of Xiaole Fan et al. [12] introduces a novel bi-branch network model named Trans-CNN Net for COVID-19 detection using CT images. This model effectively combines the strengths of Transformers and Convolutional Neural Networks (CNNs) to address the challenges posed by COVID-19 CT images, which contain both global (e.g., widespread ground-glass opacities) and local features (e.g., localized crazy-paving patterns). Traditional CNNs are adept at extracting local features but struggle with global context, whereas Transformers excel in capturing global dependencies but often require significant computational resources.

To address this, the authors designed a parallel bi-branch structure where the Transformer module extracts global features and the CNN module captures local features. These features are then fused through a bi-directional feature fusion module, enabling the model to leverage both types of information efficiently. The proposed Trans-CNN Net was evaluated on the COVIDx-CT dataset, achieving an impressive **96.7%** classification accuracy, outperforming traditional models like ResNet-152 and pure Transformer models (e.g., DeiT-B). This work demonstrates the potential of hybrid deep learning approaches in medical image classification, offering a robust tool for COVID-19 diagnosis

## 3. Theoretical Background

*3.1. Arrhythmia*

The heart beats at a regular pace, between 60 to 100 beats per minute. This rhythm is known as normal sinus rhythm, having it's origin in the depolarization of the sinoatrial node. Any disruption of this rhythm is referred to as an arrhythmia, which can manifest as an isolated irregular beat, an extended pause, or a persistent irregularity throughout the patient's life [36].

- The **P wave** represents the contraction of the atria, corresponding to their depolarization. Originating from the sinoatrial node in the right atrium, the depolarization starts in the right atrium and is followed by the left atrium. Consequently, the first half of the P wave symbolizes the depolarization of the left atrium, while the second half represents that of the right atrium rhythm [2] [32].

- The **QRS** complex represents the contractions of both ventricles and corresponds to ventricular depolarization. A wider QRS complex may

be associated with a premature ventricular contraction or a ventricular rhythm[32]

- The **R wave** is the tallest peak of the QRS complex, representing the electrical stimulus as it passes through the ventricles during depolarization. A reduced R wave progression has several causes, including an anteroseptal myocardial infarction, left ventricular hypertrophy, incorrect lead placement, etc. [46].

- The **T wave** represents ventricular repolarization. Its morphology is highly susceptible to both cardiac and non-cardiac influences, such as hormonal imbalances, certain neurological factors, myocarditis, pericarditis, fever, infections, anemia, etc. [4].

This paper aims to classify five types of cardiac arrhythmia, including atrial fibrillation (AF), first-degree atrioventricular block (IAVB), sinus bradycardia (SB), normal sinus rhythm (NSR), and sinus tachycardia (STach). These classifications are essential for improving the diagnosis and management of heart conditions, thereby contributing to the prevention of severe complications associated with these types of arrhythmia.

*Atrial Fibrillation*. During atrial fibrillation episodes, atrial activity is completely chaotic (as illustrated in Figure 3), and the atrioventricular node can be overwhelmed with over 500 impulses per minute! True P waves cannot be observed. Instead, the baseline appears flat or slightly wavy. The atrioventricular node, faced with this extraordinary surge of atrial impulses, allows only a few impulses to pass at variable intervals, resulting in an irregular ventricular rate, usually between 120 and 180 beats per minute [36].

The irregular aspect of the QRS complexes and the absence of distinct P waves is the key to identifying atrial fibrillation. The **wavy** forms that can be observed upon close inspection of the undulating baseline are called fibrillation waves [36].

Atrial fibrillation is the most common and clinically significant sustained arrhythmia in the general population. Atrial fibrillation can cause palpitations, chest pain, shortness of breath, or dizziness. A significant number of patients, especially the elderly, may have no symptoms at all and may be unaware that they are in atrial fibrillation.

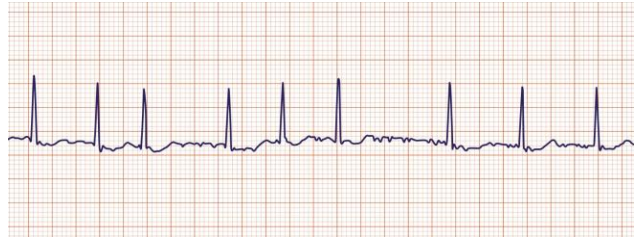
**Figure 3:** Atrial fibrillation ECG

***First-degree atriventricular block***. First-degree atrioventricular block is characterized by a delay in conduction at the level of the atrioventricular node or the His bundle, as illustrated in Figure 4. The depolarization wave spreads normally from the sinoatrial node to the atria, but when it reaches the atrioventricular node, it is delayed by more than the usual **0.1 seconds**. As a result, the PR interval (the time from the beginning of atrial depolarization to the beginning of ventricular depolarization, which includes the delay at the atrioventricular node) is **prolonged** [36].

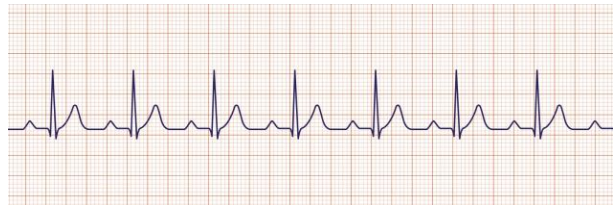
**Figure 4:** First degree atrioventricular block ECG

The diagnosis requires that the PR interval be greater than **0.2 seconds**. Despite the delayed arrival at the atrioventricular node or the His bundle, each atrial impulse eventually passes through the atrioventricular node to activate the ventricles. Each QRS complex is preceded by a single P wave [36].

***Arrhythmias of sinus origin***. **Normal sinus rhythm** (Figure 5a) is the heart's normal rhythm. Depolarization originates in the sinoatrial node. If the rhythm accelerates above 100 beats per minute, it is called sinus **tachycardia**; if it slows down below 60 beats per minute, it is called sinus **bradycardia**.

Sinus bradycardia (Figure 5c) and sinus tachycardia (Figure 5b) can be normal or pathological. Intense exercise, for example, can accelerate the

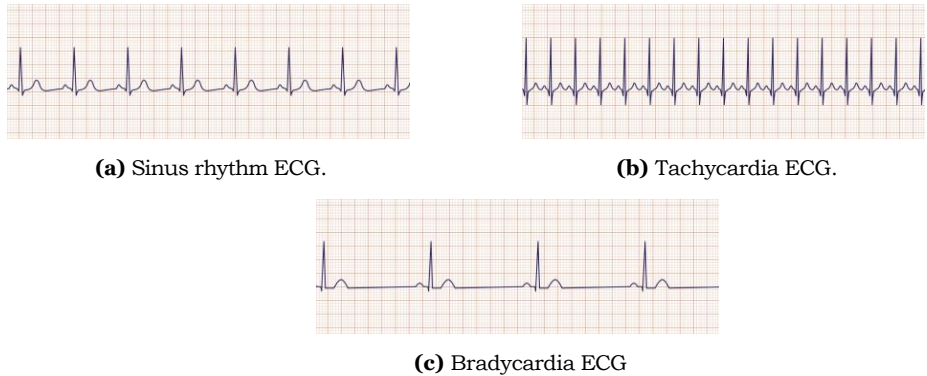

**(a)** Sinus rhythm ECG.  **(b)** Tachycardia ECG.

**(c)** Bradycardia ECG
.

**Figure 5:** Arrhythmias of sinus origin

heart rate well above 100 beats per minute, while resting rates below 60 beats per minute are typical in well-trained athletes. On the other hand, changes in the firing rate of the sinoatrial node may accompany significant heart conditions [36].

Sinus tachycardia may occur in patients with congestive heart failure or severe lung diseases, or it could be the sole initial sign of hyperthyroidism in the elderly. Sinus bradycardia can result from medications, most commonly beta-blockers, calcium channel blockers, and opioids, and it is the most frequently observed cardiac rhythm disturbance during the early stages of an acute myocardial infarction. In otherwise healthy individuals, it can arise from increased vagal tone and may lead to fainting [36].

### 3.2. Pan-Tompkins Algorithm

The Pan-Tompkins algorithm[29] is widely used for QRS detection due to its efficiency in enhancing the signal-to-noise ratio and accurately identifying QRS complexes in real time, even in noisy ECG recordings. Its adaptive thresholding and differentiation techniques make it a robust choice for various physiological conditions. Recent advancements, such as the Pan-Tompkins++ algorithm [17], have further improved its performance by introducing additional filtering and adaptive thresholding methods, enhancing accuracy in noisy environments.

### 3.3. Deep learning architectures

***Convolutional Neural Networks (CNN)***. As a subset of the deep learning algorithms, the Convolutional Neural Networks (CNN) were first used in image processing tasks. Their name is inspired from Latin (convolvere = "to roll together") and mathematics (the convolution is an integral measurement of how two functions' shapes change as they interact). A CNN architecture is a sequence of succesive groups of:

- **convolutional layer** - responsible for computing the product of two matrixes: one (the kernel) represents the set of learnable parameters of the network, the other expresses the restricted portion of the processed data. The output is a feature map, a matrix with fewer features than the original input.

- **pooling layer** - extracts the dominant characteristics of the feature map (thus reducing even more the complexity of the problem), flattens the output and passes it to the a **fully connected layer**.

To the above described sequence, the backpropagation algorithm is applied and the system is trained for a certain number of epochs.

***The Long-Short Term Memory (LSTM)***. The recurrent neural network receive as input not only the current state, but also part of information seen at a previous step. The Long-Short Term Memory cell (LSTM) [16] uses a three-gated mechanism in order to chose what information from the past to be used in the nest steps. The input gate selects the information to be stored for the current step, the unnecessary content is discarded by the forget gate, while the output gate will provide the activation to the final output of the LSTM cell. Figure 14 illustrates the structure of a LSTM cell.

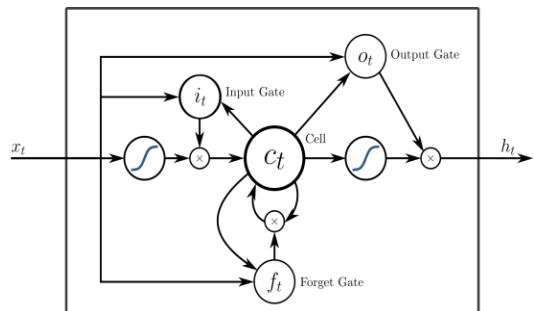

**Figure 6:** The LSTM cell [14]

***Gated Recurrent Unit (GRU)***. GRU was introduced by Kyunghyun Cho et al [6] in 2014 for Statistical Machine Translation. This layer can process sequential data such as text processing [49], speech synthesis [42] and time-series data [44]. As the name suggests, GRU unit uses a gating mechanism in order to control at each step the content of information to be passed to the hidden states. There are two types of gates: the reset gate - handles the amount of information from previous hidden state which should be forgotten and the update gate - handles how much from the new input should be used to update the hidden state The output of the GRU is calculated based on the updated hidden state. A GRU cell is illustrated in Figure 12.

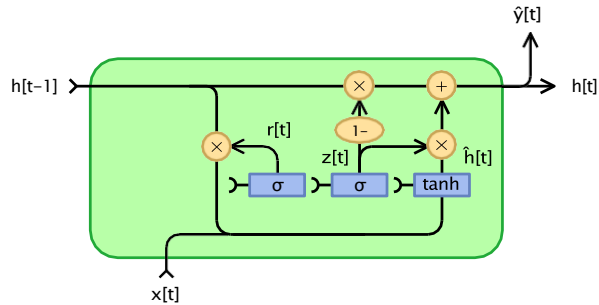

**Figure 7:** GRU unit [19]

***Vision Transformers (ViT)***. Although initially applied in machine translation, tranformers [40] achieved excellent results in the field of image classification [9]. The transformer architecture consists in a sequence-to-sequence approach enriched with an attention mechanism, which was designed to emphasize relationships between data regardless of their position in the sequence (in a past step or in a future one).

***Residual Networks - ResNet***. are a deep learning architecture in which the layers learn residual functions with reference to the layer inputs. It was developed in 2015 by Kaiming He et al.[15] for image recognition, and won the ImageNet Large Scale Visual Recognition Challenge (ILSVRC) of that year[3]. Within the residual blocks, the input of the block is directly added to its output, forming a residual connection (or a skip connection), as illustrated in Figure 8. This architectural innovation allows for the efficient training of very deep networks by alleviating the vanishing gradient problem and facilitating the learning of identity mappings.

---

[3]https://image-net.org/challenges/LSVRC/2015/results.php

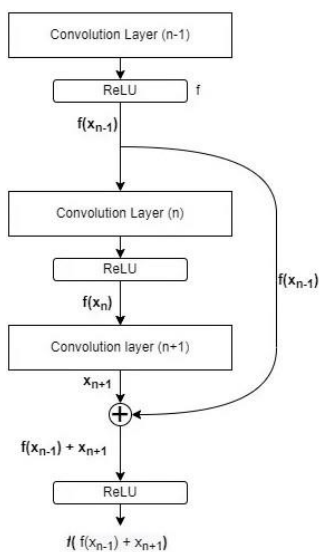

**Figure 8:** A skip connection within ResNet

The fundamental distinction between ViTs and ResNet lies in their architectural frameworks. ViTs leverage a self-attention mechanism, inspired by transformer models, which facilitates the capture of both global and local dependencies within an image. This mechanism endows ViTs with superior adaptability, few-shot learning capabilities, and scalability, making them particularly well-suited for tasks involving limited data. In contrast, ResNet is characterized by its depth and the use of residual connections, which enable the efficient capture of hierarchical features and promote the training of very deep networks

## 4. Preprocessing

### 4.1. QRS extraction

Extracting the QRS complexes is an important step in the analysis of ECG signals, as it is essential for diagnosing arrhythmias. To detect the QRS intervals, we implemented a function based on the Pan-Tompkins alorithm [17, 29], that applies a series of preprocessing and filtering steps to isolate the QRS complexes from the ECG signal:

**Algorithm 1** Detect QRS Complexes in ECG Signal
---
**Require:** signal: 1D ECG signal array
**Require:** signal_freq: Sampling frequency (Hz)
**Ensure:** R-peaks, Q-peaks, S-peaks indices
 1: **Initialize** lowcut = 5.0, highcut = 15.0, filter_order = 2
 2: Compute Nyquist frequency: nyquist_freq = 0.5 × signal_freq
 3: Normalize cutoff frequencies: low = lowcut/nyquist_freq, high = highcut/nyquist_freq
 4: Design bandpass filter: (b, a) = butter(filter_order, [low, high], btype="band")
 5: Apply filter: y = lfilter(b, a, signal)
 6: Compute derivative: diff_y = diff(y)
 7: Square the derivative: squared_diff_y = diff_y$^2$
 8: Apply moving window integration: integrated = convolve(squared_diff_y, $\mathbf{1}_5$)
 9: Normalize: normalized = $\frac{integrated - \min(integrated)}{\max(integrated) - \min(integrated)}$
10: Detect candidate peaks: peaks = find_peaks(normalized, distance = signal_freq/5, height = 0.5, width = 0.5)
11: Initialize empty lists: R_peaks, Q_peaks, S_peaks
12: **for all** peak ∈ peaks **do**
13:     Find max in local window: r_peak = arg max signal[peak − signal_freq/5 : peak + signal_freq/5]
14:     Append r_peak to R_peaks
15: **end for**
16: **for all** r_peak ∈ R_peaks **do**
17:     Search Q peak in window before R: q_peak = arg min signal[r_peak − signal_freq/4 : r_peak]
18:     Append q_peak to Q_peaks
19:     Search S peak in window after R: s_peak = arg min signal[r_peak : r_peak + signal_freq/4]
20:     Append s_peak to S_peaks
21: **end for**
22: **return** (R_peaks, Q_peaks, S_peaks)

*4.2. Converting ECGs into images*

We apply 2D models on the signal images generated by the algorithm below. The chosen image size (506x187), maintains quality without compromising performance, ensuring faster training and classification, making it more suitable for use on hardware with limited capabilities. While Pan-Tompkins is generally reliable, it can produce false positives in the presence of noise or ectopic beats. In our case, the algorithm performs poorly in situations with extreme noise, and we did not implement post-processing specifically to handle these issues. If there are visible instances of abnormal noise that are clearly identifiable in the signal, the image will be excluded from the classification process to prevent inaccurate results. This helps maintain the integrity of the classification even in challenging conditions.

---

**Algorithm 2** Convert ECG Signals to Images

---

**Require:** ECG signal array for 12 leads
**Ensure:** Grayscale image of size $506 \times 187$
1: **Step 1: Plot ECG Signals**
2:  Plot all 12-lead ECG signals in a single figure  ▷ Use a plotting library like Matplotlib
3: **Step 2: Save the Figure**
4:  Export the plotted figure as an image file (e.g., PNG format)
5: **Step 3: Load and Process Image**
6:  Read the saved image using OpenCV
7:  Convert the image to grayscale
8:  Resize the image to $506 \times 187$ pixels
9: **Step 4: Save the Processed Image**
10: Save the final grayscale image for further analysis

---

Transforming 1D ECG signals into 2D images allows deep learning models to extract spatial and texture-based features that may be less apparent in raw waveforms. CNNs excel at detecting local patterns, such as wave morphology and textures, through hierarchical feature extraction using convolutional filters. Transformers, on the other hand, leverage self-attention to capture long-range dependencies and global contextual relationships within the image. This transformation enables the use of well-optimized vision architectures, improving pattern recognition and classification performance.

All these steps were essential to prepare the dataset for training the 2D CNN model, ensuring consistency and quality in the input data.

## 5. Experimental setup

*5.1. Dataset*

In the development of this work, the following datasets were used: PTB-XL electrocardiography dataset [41], PTB Diagnostic ECG Database [3], China 12-Lead ECG Challenge Database [24], Georgia 12-Lead ECG Challenge Database [13], and St. Petersburg INCART 12-lead Arrhythmia Database [37]. These datasets contain electrocardiograms in 12 leads. The selected arrhythmias include atrial fibrillation (AF), first-degree atrioventricular block (IAVB), sinus bradycardia (SB), normal sinus rhythm (NSR), and sinus tachycardia (STach).

All the datasets used contain .mat and .hea files recorded at a 500 Hz frequency and follow the same format. Therefore, we simply combined all the files into one location and used them without requiring data cleaning, since there were no discrepancies in the labeling standards. We selected AF, IAVB, SB, NSR, and STach because these arrhythmias are generally asymptomatic or have a minimal mortality risk, so the penalty for occasional misclassification is lower. Additionally, these classes had the most recordings available across the datasets, making them ideal for training.

The number of recordings varied between classes, so in order to balance the dataset, an equal number of examples from each selected arrhythmia was used, amounting to a total of **8360** recordings, with **1672** for each type of arrhythmia. The extra recordings from each class were discarded so that the number of recordings in each class was exactly the same, avoiding bias in the training process and ensured that no class was overrepresented. The dataset was split into 80% for training and validation, utilizing 10-fold cross-validation for model evaluation. The remaining 20% of the dataset was reserved for testing, ensuring an unbiased assessment of the model's performance.

For our experiments, we analyzed the input data from two perspectives. First, we fed the entire ECG signal into a raw 1D CNN-based architecture. Building on this foundation, we developed three additional variants of the model by incorporating the QRS complexes: a 1D CNN + GRU model, a GRU-based model, an LSTM-based model, and a GRU + LSTM hybrid model. Notably, even the GRU-based, LSTM-based, and GRU + LSTM models included a 1D CNN layer as a preliminary step. This inclusion helped accelerate model training and improve performance by enabling the model to extract key temporal features from the ECG data efficiently.

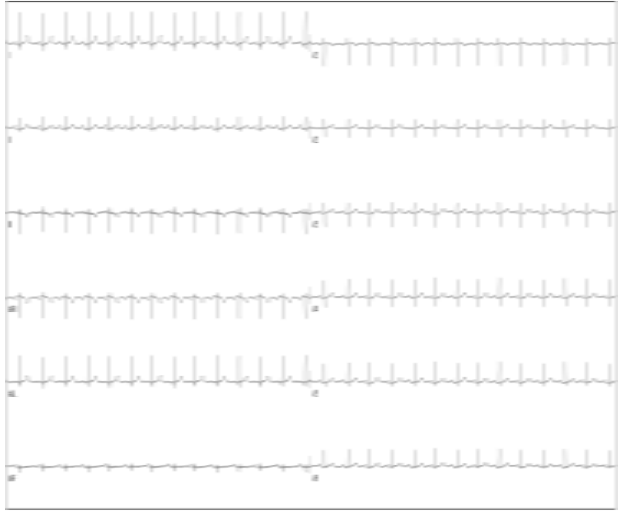

**Figure 9:** Input data for the CNN 2D architecture

Additionally, we designed one more model using a 2D CNN, where the ECG signal was converted into images as input. Furthermore, starting from the related work described in Section 2, we experimented with several external models to test their performance on the image-based data. These included CNN-based models like ResNet and transformer-based architectures such as untrained ViT and Deep ViT, as well as pretrained vision transformers, including ViT, DeiT, and BeiT. The code[4] was implemented in Python using the PyTorch library[5] and ran on a system with GPUs (NVIDIA's GeForce GTX 1080 Ti, 11GB memory).

Each model was rigorously evaluated using a 10-fold cross-validation approach to ensure robust and reliable performance metrics. To optimize the training process, we employed techniques such as reducing the learning rate on plateau and early stopping. These strategies helped prevent overfitting and ensured convergence to the best-performing model configurations.

---

[4]The code is available on private email request.
[5]https://pytorch.org

## 6. Proposed Models

In the first scenario, the proposed models employ a one-dimensional convolutional neural network (1D CNN) for ECG arrhythmia classification (Figure 10). Aiming to improve the system accuracy, we enriched the convolutional layers with different combinations of LSTM and GRU layers, resulting four more architectures described in Figures 11-14. All the systems were trained with the **CategoricalCrossentropy** loss function and the **Adam** optimizer, starting with a learning rate of 0.001, which is progressively reduced to 0.0000016. A batch size of 50 and 50 epochs are used for training. In terms of evaluation, the 10-fold cross-validation was used to ensure robust performance assessment.

The input data comprises 8360 ECG signals with 12 leads, divided into 1672 samples per class for each of the 5 arrhythmia classes, ensuring balanced training and evaluation, as described in Section 5. The hyperparameters and the architectures of each model are synthesised in Table 2. All the paramters were chosen based on initial tests.

**Table 2:** Systems' parameters and architectures

| System | Convolutional Block | | | Hidden Layer |
|---|---|---|---|---|
| | no of kernels | size of kernels | no of filters | |
| 1D CNN | 3 | 8,5,3 | 64 each | N/A |
| 1D CNN+GRU | 3 | 8,5,3 | 64 each | 128 GRU units |
| GRU | 3 | 8,5,3 | 64 each | 64, 128 GRU units |
| GRU+LSTM | 3 | 8,5,3 | 128 each 2*128 LSTM units | 2*128 GRU units+ |
| LSTM | 3 | 8,5,3 | 128 each 2*256 LSTM units | 128 LSTM units+ |

***Backbones Architectures (Trained from scratch)***. To classify the arrhythmia based on the image of the ECG signal, we started with the CNN architecture described previously in Table 2.

Then, we annalysed the backbones CNN architecture in the field of image classification, ResNet [15], ViT [9] and DeepVit [48] and we trained each system from scratch for 300 epochs over our dataset. The results are described in Table 4.

***Pretrained Backbone Architectures***. Inspired by the works synthesised in [25] and [34], we started from the pretrained ViT [9], BEiT [1], DeiT [38] and we fine-tuned each system for 10 epochs over our dataset, with an early stopping mechanism of 3 epochs patience step. The results are described in Table 4.

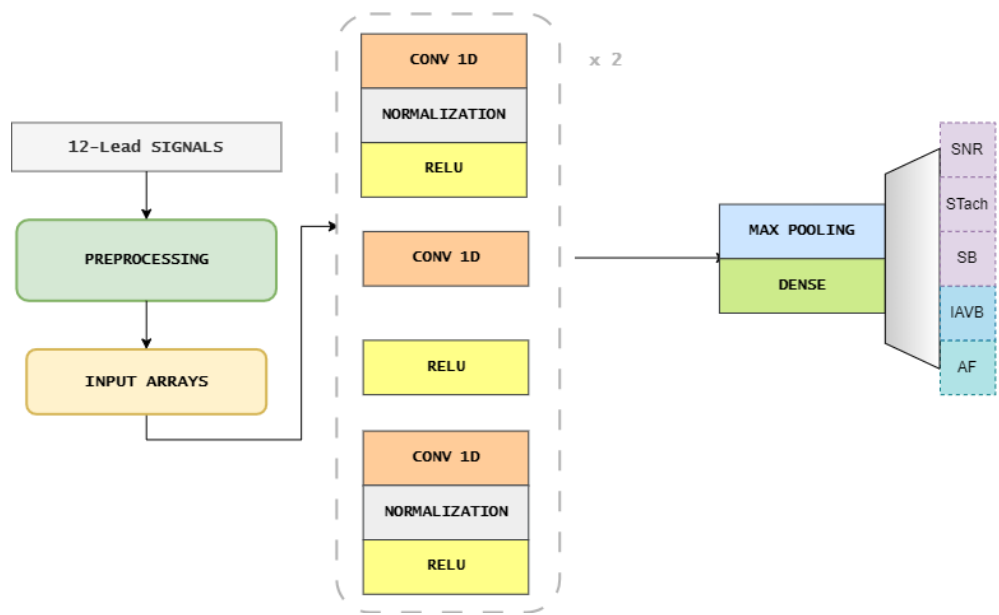

**Figure 10: 1D CNN Model** The proposed model, inspired by the work of Pranav Rajpurkar and Awni Y. Hannun [31], as well as Madhura Deshmane and Swati Madhe [7], employs a one-dimensional convolutional neural network (1D CNN) for ECG arrhythmia classification.

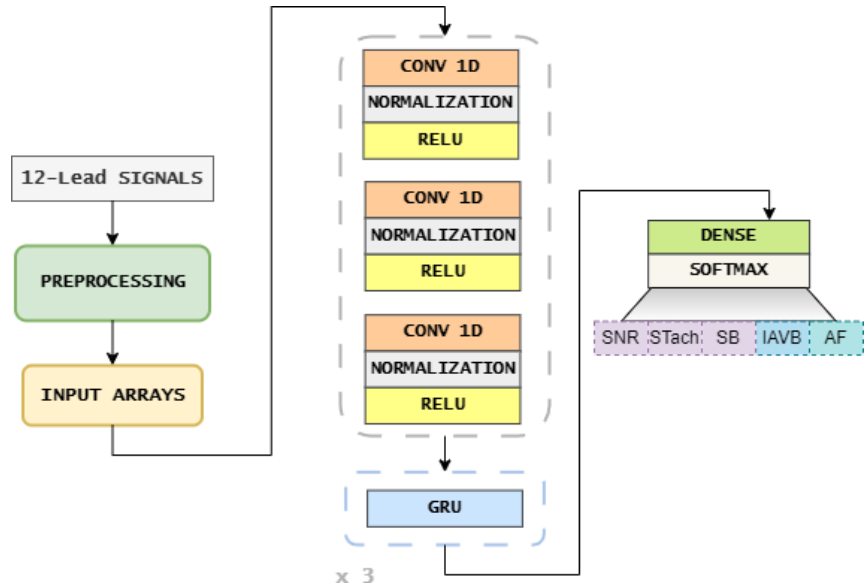

**Figure 11: 1D CNN with GRU** This model combines a 1D CNN with a Gated Recurrent Unit (GRU) for ECG arrhythmia classification, inspired by the research of Tsai-Min Chen [5].

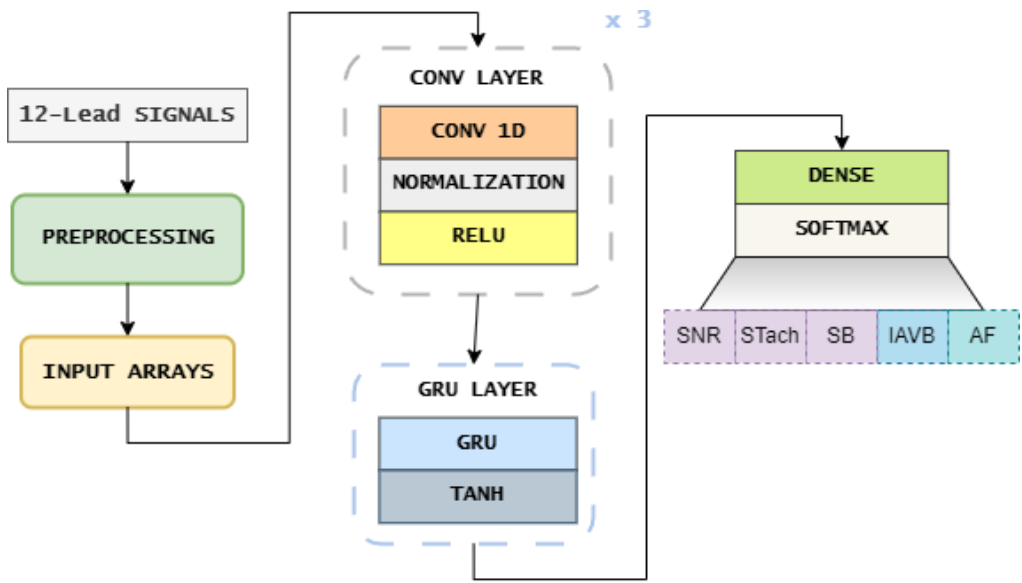

**Figure 12: GRU Model** This model uses a Gated Recurrent Unit (GRU)-based architecture for ECG arrhythmia classification, inspired by the works of Weiran Ji [20] and Zhaohan Xiong [43].

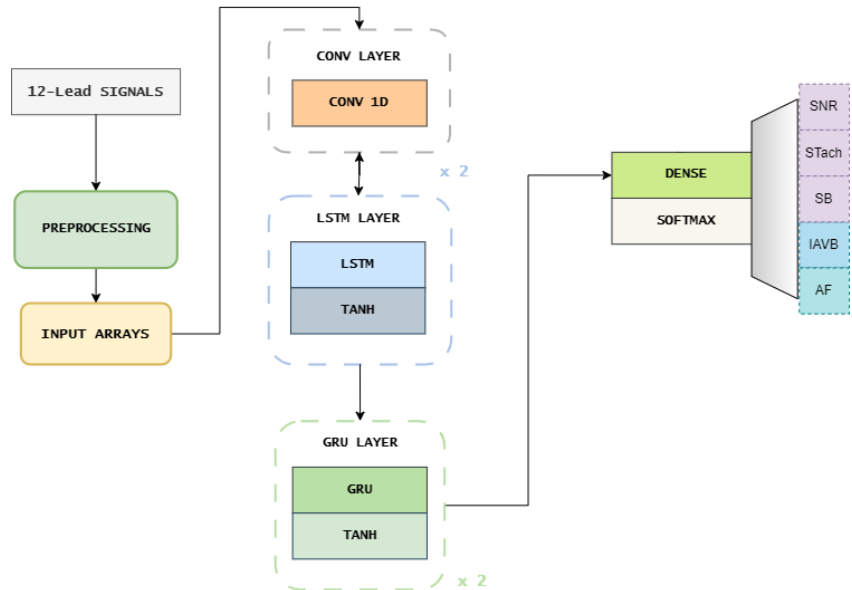

**Figure 13: GRU+LSTM Model** This model combines GRU and LSTM units for ECG arrhythmia classification. Similar to the previous models.

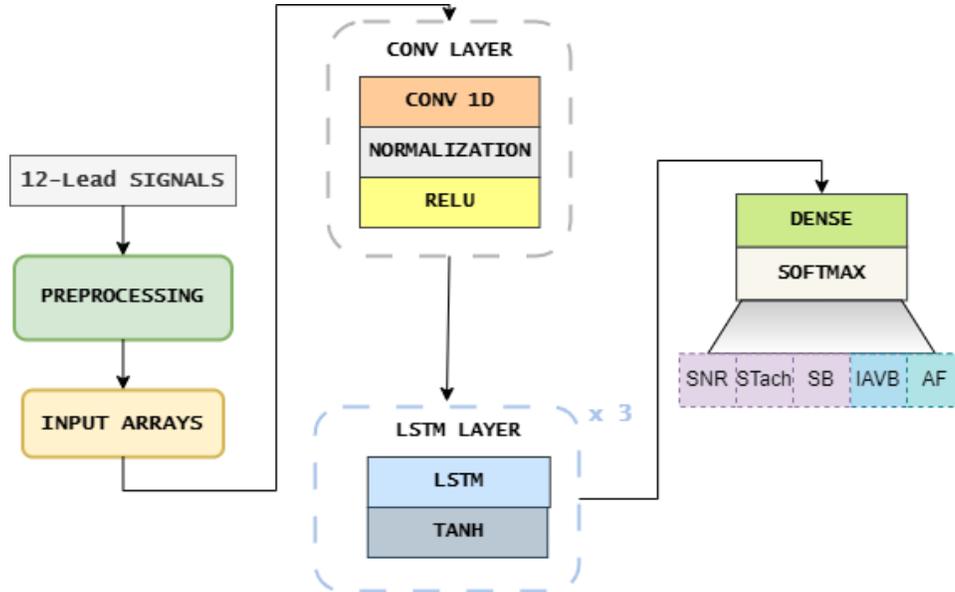

**Figure 14: LSTM Model** This model employs LSTM layers for ECG arrhythmia classification. A similar approach is described in "ECG-Based Arrhythmia Classification using Recurrent Neural Networks in Embedded Systems" [11].

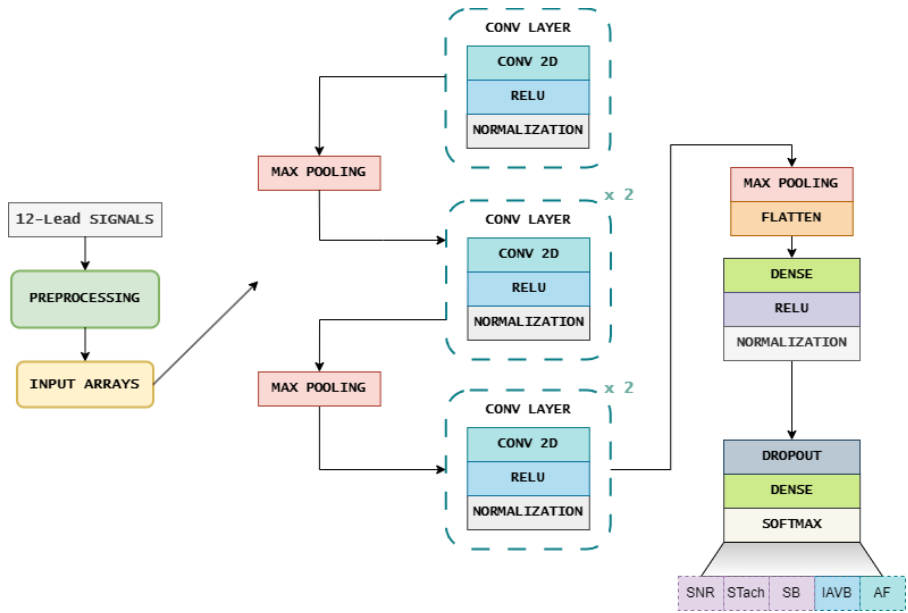

**Figure 15: 2D CNN Model** This model utilizes a two-dimensional convolutional neural network (2D CNN) for ECG arrhythmia classification, drawing on the methodology of Ullah Amin and Rehman Sadaqat [39].

## 7. Results and Discussions

All 12 systems were evaluated with the *classification accuracy metrics*, which is expressed as the ratio between the correct predicted items out of the total samples.

Given the medical context, we have also considered the *F1 score*, which is the harmonic mean of precision and recall, meaning that it penalizes extreme values of either, as a low precision combined with a high recall implies that the classifier is prone to set of false alarms.

$$F1 = 2 \times \frac{\text{precision} \times \text{recall}}{\text{precision} + \text{recall}}$$

To be more specific, according to the Ntional Library of Medicine[6], the precision denotes the proportion of the retrieved samples which are relevant and is calculated as the ratio between correctly classified samples and all samples assigned to that class. The recall denotes the rate of positive samples correctly classified, and is calculated as the ratio between correctly classified positive samples and all samples assigned to the positive class.

---

[6]https://tinyurl.com/PubMedCentraLinkl

The results for the CNN1D architectures are illustrated in Table 3 (on validation datset) and Figure 16 (for both train and validation steps). We can notice that the GRU based architecture achieved the highest accuracy rate.

| System ID | Accuracy | F1 score |
|---|---|---|
| GRU+LSTM | 66.98% | 0.6589 |
| LSTM | 83.97% | 0.8380 |
| CNN1D | 90.78% | 0.9075 |
| GRU | 91.62% | 0.9140 |
| **CNN1D+GRU** | **93.42%** | **0.9338** |

**Table 3:** Accuracy and F1 score for the 1D systems fed with the ECG signals, computed on the validation data.

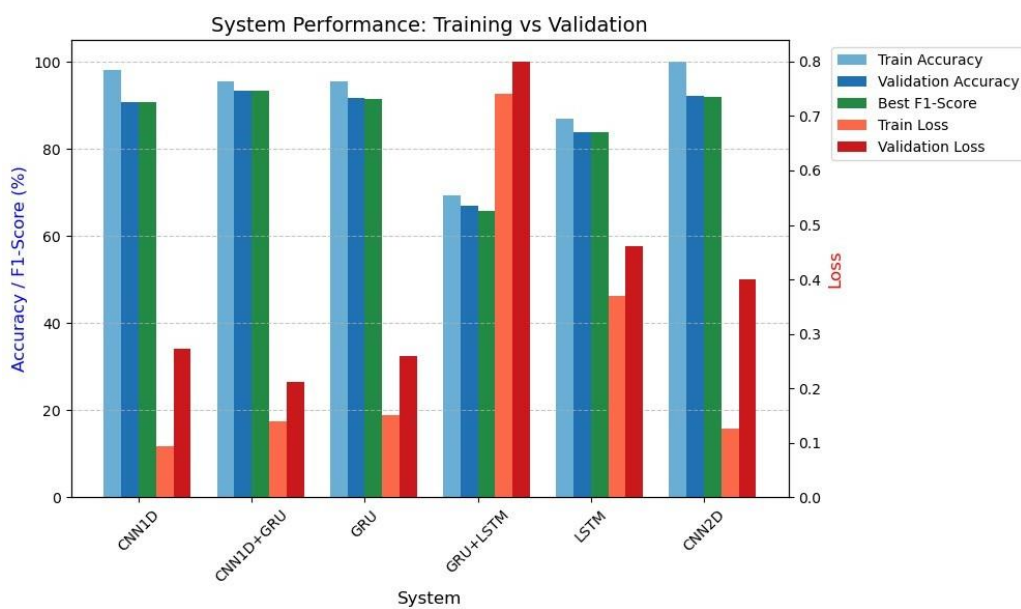

**Figure 16:** Systems' performances in terms of accuracy and loss when trained using the 12-Lead ECG Signals

From Table 4 we can conclude that the hybrid system CNN1D+GRU

obtained the highest accuracy of 93.42%, along with the highest F1-score 0.9338.

| System ID | Accuracy | F1 score |
|---|---|---|
| **CNN2D** | **92.16%** | **0.9194** |
| ResNet_untrained | 90.66% | 0.91 |
| ViT_untrained | 73.49% | 0.72 |
| DeepVit_untrained | 70.98% | 0.69 |
| ViT_pretrained | 84.60% | 0.84 |
| BeiT_pretrained | 81.89% | 0.79 |
| DeiT_pretrained | 84.60% | 0.81 |

**Table 4:** Accuracy and F1 score for the 2D systems fed with the ECG images, computed on the validation data.

The confusion matrices from Figure 17 show that 1st degree Atrioventricular Block (IAVB) exhibits the poorest classification performance, evidenced by numerous errors categorizing it as other types. This lower accuracy is likely due to IAVB's subtle characteristics, such as a prolonged PR interval, which can resemble normal sinus rhythms (SNR) or sinus bradycardia (SB). The significant misclassification into SB and SNR suggests that the network struggles to differentiate IAVB from these rhythms, probably due to overlapping features or a lack of training data highlighting the significance of the PR interval. Furthermore, the moderate misclassification of IAVB as AF or STach suggests more general problems with feature representation, which could be mitigated through improved feature extraction techniques.

Our classification models can be integrated into clinical workflows by embedding them into decision-support systems that receive ECG signals and provide automated arrhythmia predictions. In a cardiologist's diagnosis pipeline, these predictions could serve as an initial screening tool, flagging abnormal rhythms for further review and aiding in triage. The cardiologist can then analyze the raw ECG data alongside the model's insights to confirm or refine the diagnosis. Additionally, these models have educational applications, helping medical students learn ECG interpretation by visualizing arrhythmia patterns, QRS complexes, and other key features, fostering a deeper understanding of electrophysiological signals.

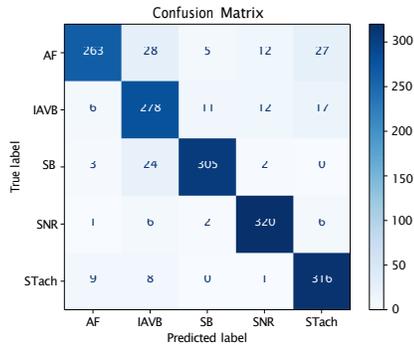
(a) ResNet Model

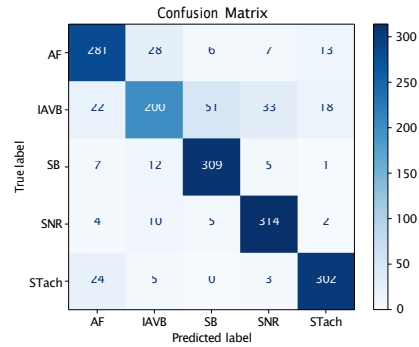
(b) ViT Model Pretrained

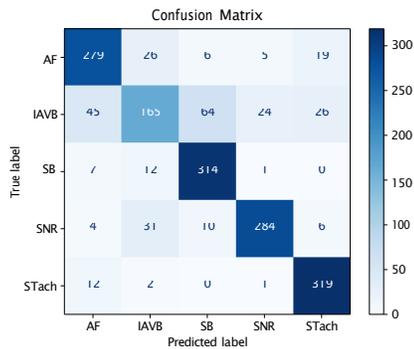
(c) BeIT Model Pretrained

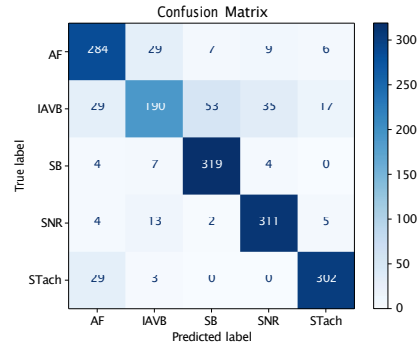
(d) DeIT Model Pretrained

**Figure 17:** Confusion matrixes for our best CNN2D models, computed on the dataset reserved for testing.

### 7.1. Comparison with State-of-the-Art Benchmarks

When comparing our results with existing literature, we observe that many prior works achieve exceptionally high accuracy, particularly those focusing on beat classification rather than direct arrhythmia classification. Approaches such as those by Tae Joon Jun et al.[21] and Dhyani et al.[8] report accuracy values as high as 99%, but they classify individual heartbeats rather than full arrhythmia episodes. While this approach is effective, it requires an additional post-processing step to group beats into arrhythmia categories, which can introduce complexity and potential misclassification when applied in real-world clinical settings. However, studies that classify arrhythmias directly, such as those by Zhang et al.[47] and Li et al[23], of-ten employ deep architectures like CNNs with SHAP analysis or specialized networks such as DSE-ResNet. Although these models achieve excellent accuracy, their complexity can pose challenges for real-world deployment, particularly in resource-limited medical facilities that may not have the computational infrastructure to support such models efficiently. Our results, while not surpassing the top-performing methods, demonstrate competitive performance with a lightweight approach. The best-performing model in our study, CNN1D + GRU, achieves 93.42% accuracy, which is higher than several related works that rely on simpler deep learning architectures. Similarly, our CNN2D model, trained on ECG images, achieves 92.16% accuracy, making it a strong contender in 2D-based classification.

## 8. Conclusions and Future Work

In this paper we have analysed 12 different deep convolutional neural and transformer based networks in the context of arrhythmia classification. The systems were trained on labeled pairs of ECG images/signals and the corresponding type of arrhythmia. The input data was processed within two scenarios: the exracted QRS complexes and the ECG images.

While related works often achieve exceptional results by focusing on individual heartbeat classification, using inputs such as segmented beats, our approach diverges by leveraging the entire ECG signal in various ways to classify arrhythmias. This comprehensive analysis of the full ECG signal allows us to capture broader temporal and morphological patterns that may span multiple heartbeats, offering a holistic perspective. In contrast, other works primarily classify arrhythmias by identifying and analyzing each abnormal beat in isolation, which, while effective for beat-specific abnormalities, might miss the contextual information provided by the continuous signal.

Enhancing our feature extraction and preprocessing steps could further refine classification performance, particularly in differentiating challenging arrhythmia types. Future work should focus on optimizing feature engineering while keeping computational efficiency in mind, ensuring that our approach remains lightweight and accessible for lower-income regions where high-end computational resources may not be available. By leveraging advancements in machine learning, we plan to incorporate techniques such as deep learning and feature extraction to adapt the algorithm to various signal variations and noise levels.

In terms of real-world applicability, our approach could be integrated into wearable ECG monitoring systems for real-time arrhythmia detection. A patient fitted with a wearable device, such as a smartwatch or chest patch, would have their ECG continuously recorded and transmitted either to a local processing unit (e.g., a smartphone) or a cloud-based server. The classification model would then analyze the signal in real time, detecting potential arrhythmias. If an abnormal rhythm is identified, alerts could be sent to both the patient and their doctor, allowing for timely medical intervention.